\documentclass[11pt]{article}

\usepackage{acl}
\usepackage{times}
\usepackage{latexsym}
\usepackage[T1]{fontenc}
\usepackage[utf8]{inputenc}
\usepackage{microtype}
\usepackage{inconsolata}
\usepackage{graphicx}
\usepackage{booktabs}
\usepackage{tikz}
\usepackage{fvextra}
\usetikzlibrary{shapes, arrows.meta, positioning}
\title{LLM-Assisted Stance Detection in Scientific Discourse: A Test Case in Bayesian Cognitive Science}

\author{
  Eyup Engin Kucuk\thanks{Corresponding author: \texttt{eyup.kucuk@unh.edu}. } \\
  University of New Hampshire \\
  Department of Psychology
  \And
  Tarik Kelestemur \\
  Independent Researcher
  \And
  Ömer Dağlar Tanrikulu \\
  University of New Hampshire \\
  Department of Psychology
}

\begin{document}
\maketitle

\begin{abstract}
Qualitative coding is central to social science, but expert annotation is difficult to scale.
LLMs offer a possible extension, yet require careful validation when the target construct is interpretive, theoretically loaded, and only indirectly expressed.
We study this problem in a difficult case: detecting whether authors treat Bayesian models as descriptions of mental and neural mechanisms (realism) or as useful mathematical tools (instrumentalism).
Our method combines a theory-driven codebook, expert-coded reference annotations, a diagnostic-gated prompt-optimization search yielding a shared zero-shot prompt for three frontier LLMs (GPT-5.1, Claude Sonnet 4.6, Gemini 3 Pro Preview), and multi-rater reliability analysis.
The final prompt achieved a held-out combined reliability score of 0.76 (harmonic mean of ICC = 0.79 and $\alpha$ = 0.74), with all diagnostics satisfied.
Deployed on 6{,}858 quotes from 210 articles, the three LLMs reached substantial quote-level agreement (ICC = 0.80; $\alpha$ = 0.76; combined = 0.78) and near-perfect article-level rank stability ($r$ = 0.96--0.97 across rater pairs).
The corpus was predominantly weakly realist, but article-level stances were rarely uniform: only 1.4\% of articles used a single band, while 59.5\% spanned four or more.
Low-level perception/motor articles scored 8.8 Realism points higher than high-level cognition articles ($p{<}.001$, $d{=}0.60$), quantifying a long-held qualitative intuition.
We present this as an expert-led case study; the framework is intended to generalize to similar theoretically demanding tasks, not to all qualitative analysis.\footnote{Code, data, and reference annotations: \url{https://github.com/EyupEK/autoresearch_bayes}}
\end{abstract}

\section{Introduction}

Qualitative coding is central to social science, but its interpretive depth limits scale.
This is especially true when the target is not a surface topic but a theoretically meaningful stance, frame, or commitment that the author does not explicitly name.
LLMs offer a possible route past this constraint, and recent work has evaluated them as annotators for political text, qualitative coding, content analysis, and scalar constructs \citep{gilardi2023,tornberg2024,chew2023,dunivin2024,dunivin2025,licht2025,than2025,ziems2024}.
The central problem is how to know whether model outputs are reliable enough to support qualitative inference: apparent agreement can be produced by shallow shortcuts such as defaulting to a frequent class or avoiding fine-grained scale use, and the question is sharper when the construct is theoretical and not surface-realized in any single phrase.

We examine this problem in Bayesian cognitive science, where authors often make claims relevant to the realism--instrumentalism distinction without naming either position.
Some passages treat Bayesian models as useful predictive tools; others imply that Bayesian computations correspond to mental or neural mechanisms.
These distinctions are philosophically deep \citep{jones2011,bowers2012,griffiths2012}, often hedged, and domain-dependent.

This paper makes \textbf{three contributions}.
(1)~We contribute a stance-detection method in qualitative analysis employing an automated, diagnostic-aware prompt-optimization loop that delivers a deployable annotation prompt.
(2)~We showcase how to apply the autoresearch method to social science research in order to optimize a single shared prompt across three heterogeneous frontier LLMs, treating the prompt as a common annotation protocol rather than per-model tuning.
(3)~We contribute a codebook design pattern for operationalizing theoretically complex and indirectly expressed social science concepts through the example use case of a working codebook for realism vs.\ instrumentalism in Bayesian cognitive science.

The framework's components (theory-driven codebook architecture, diagnostic-aware prompt optimization, and shared-prompt multi-rater convergence) are designed to be portable to similar theoretically demanding qualitative coding tasks, though the specific codebook and prompt must be adapted to each new construct; cross-domain validation is intended future work.

\section{Related Work}

\paragraph{LLM-as-annotator for computational social science.}
A growing literature evaluates frontier LLMs as text annotators for social-science tasks.
\citet{gilardi2023} report that GPT-3.5 matches or exceeds crowdworker accuracy on five English-language political and content-moderation tasks; \citet{tornberg2024} extend this to ideological classification.
\citet{ziems2024} provide the largest evaluation to date (13 LLMs on 25 CSS benchmarks) and conclude that zero-shot LLMs reach fair agreement with humans on classification, though they fail to match fine-tuned models.
Recent work flags substantial variability across tasks, prompts, models, and re-runs \citep{kristensen2025}, with \citet{camuffo2026} cataloguing five distinct sources of variance in LLM annotation (construct specification, interface effects, model preferences, output extraction, and system-level aggregation) and arguing for variance-aware protocols rather than ad hoc deployment.
Closer to our setting, \citet{chew2023,dunivin2024,dunivin2025,than2025,gao2024,meng2024} have studied codebook-based or theory-driven LLM workflows in qualitative coding; our task, detecting an implicit philosophy of science stance, sits at the harder end of this literature \citep{simons2026} because the construct is not surface-realized in any single phrase.

\paragraph{Reliability and stance detection.}
On the evaluation side, our reliability analysis builds on the long-standing literature on intraclass correlation \citep{shrout1979}, Krippendorff's $\alpha$ for ordinal annotations \citep{krippendorff2011}, and significance testing in NLP \citep{koehn2004,berg-kirkpatrick2012,dror2018}.
Recent work has begun separating signal from systematic noise in annotator disagreement \citep{ivey2025nutmeg}; we contribute a complementary angle by decomposing reliability across analytical grain sizes.
Standard metrics summarize agreement at the item level, but when downstream analyses operate on aggregates (documents, authors, subfields), per-item agreement can understate or overstate the agreement that actually matters.
We complement quote-level $\alpha$ with article-level rank stability via cross-rater correlations of per-article random intercepts from per-rater linear mixed-effects models.
Stance detection has traditionally focused on expressed positions toward named targets \citep{burnham2025}; here the ``target'' is itself a philosophical position the author rarely names.

\paragraph{Realism and instrumentalism in Bayesian cognitive science.}
Whether Bayesian models describe what minds and brains actually do, or are useful mathematical idealizations, has been argued explicitly in the field \citep{danks2008,jones2011,bowers2012,griffiths2012,rescorla2023,rescorla2025}.
Two recent positions directly shape our annotation design: \citet{williams2023} argue the realism--instrumentalism dichotomy is misleading when treated as a forced choice between mechanistic and black-box modelling, and \citet{rescorla2023,rescorla2025} emphasize that both are graded stances, since a researcher can be realist about some aspects of the framework while instrumentalist about others.
Our codebook accommodates this graded view via complementary continuous scales constrained to sum to 100, paired with ordinal bands; where the philosophical literature works with hand-picked quotations, we operationalize this construct at corpus scale.

\begin{figure*}[t]
\centering
\begin{tikzpicture}[
  node distance=0.45cm and 0.6cm,
  font=\footnotesize,
  every node/.style={align=center},
  expert/.style={rectangle, rounded corners=3pt, draw=black!70, fill=blue!20,
                 minimum width=2.15cm, minimum height=0.95cm, inner sep=3pt},
  auto/.style  ={rectangle, rounded corners=3pt, draw=black!70, fill=orange!25,
                 minimum width=2.15cm, minimum height=0.95cm, inner sep=3pt},
  deploy/.style={rectangle, rounded corners=3pt, draw=black!70, fill=green!22,
                 minimum width=2.15cm, minimum height=0.95cm, inner sep=3pt},
  arr/.style={-Latex, thick, draw=black!70},
  larr/.style={-Latex, thick, draw=black!70, dashed},
]
\node[expert]                    (codebook) {Theory-driven\\codebook};
\node[expert, right=of codebook] (ref)      {251 expert-coded\\quotes\\\scriptsize(151 / 50 / 50)};
\node[auto,   right=of ref]      (loop)     {Autoresearch\\prompt search\\\scriptsize(propose $\to$ score $\to$ gate)};
\node[auto,   right=of loop]     (final)    {Final shared\\zero-shot prompt};
\node[deploy, right=of final]    (rate)     {GPT-5.1, Claude,\\Gemini annotate\\6{,}858 quotes};
\node[deploy, right=of rate]     (out)      {Multi-level\\reliability +\\corpus analysis};

\draw[arr] (codebook) -- (ref);
\draw[arr] (ref)      -- (loop);
\draw[arr] (loop)     -- (final);
\draw[arr] (final)    -- (rate);
\draw[arr] (rate)     -- (out);

\draw[larr] (loop.north) .. controls +(0.55,0.9) and +(-0.55,0.9) ..
            node[above, font=\scriptsize, yshift=-2pt] {revise on reject} (loop.north);
\end{tikzpicture}
\caption{Study workflow. \textbf{Purple}: expert-defined inputs. \textbf{Orange}: automated prompt optimization (detailed in Figure~\ref{fig:autoresearch}). \textbf{Green}: deployment and multi-level reliability analysis.}
\label{fig:workflow}
\end{figure*}
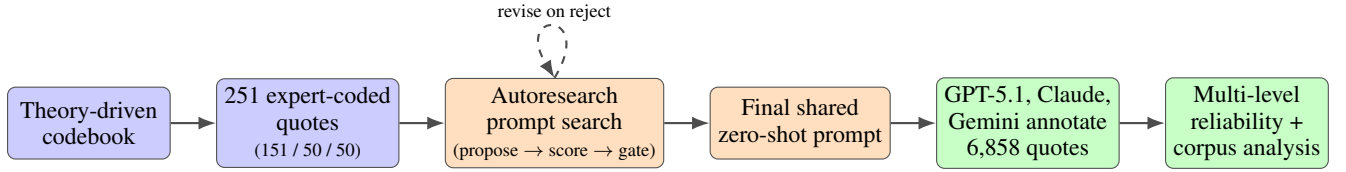

\section{Dataset and Codebook}
\label{sec:data}

\subsection{Codebook}

The codebook defines complementary Realism and Instrumentalism scores constrained to sum to 100, each on a seven-band ordinal scale.
Realism bands span Anti-Realist (0--10), Weak Realist (11--27), Realism-Sceptical (28--44), Agnostic (45--55), Leaning Realist (56--72), Moderately Realist (73--89), and Strong Realism (90--100); Instrumentalism mirrors these ranges as the complementary pair.
Within each assigned band, raters choose the midpoint or midpoint $\pm$4 or $\pm$8 to register the strength of the band-defining cues.
Each band combines a semantic definition with linguistic-criteria tags for Subject (e.g., model / agent / mechanism), Claim Target (function / mapping / mechanism), Modality (neutral / hedged / strong / as-if), and Evidence Link (none / inferred / conceptual / direct), enabling both interpretability and automated validation of the model's chosen tags against codebook-allowed tags per band.
The full Realism scale, including per-band Upper-Band and Lower-Band specifications for each axis, is reproduced in Appendix~\ref{app:codebook} (Figure~\ref{fig:codebook}); the Instrumentalism scale mirrors this structure.

\subsection{Reference Annotations}

One domain expert read 15 articles in Bayesian cognitive science and extracted 251 \emph{assumption-bearing quotes}, passages making an implicit or explicit claim about Bayesian modeling as an explanation, description, or prediction of a behavioral, functional, cognitive, or neural capacity (Table~\ref{tab:abq}).
The same expert annotated each quote with the codebook.
The 251 quotes were partitioned into a fixed 151 / 50 / 50 tuning / development / held-out validation split.
We call the first partition a tuning set because no model parameters are updated; it is used only to evaluate candidate prompts.
These labels are expert-coded reference annotations rather than multi-expert adjudicated ground truth.

\begin{table}[t]
\centering
\small
\setlength{\tabcolsep}{6pt}
\renewcommand{\arraystretch}{1.15}
\begin{tabular}{@{}p{0.55\columnwidth}p{0.40\columnwidth}@{}}
\toprule
\textbf{Quote} & \textbf{What it claims} \\
\midrule
\textit{``These results provide stronger evidence for the hypothesis that the brain computes with sensory uncertainty.''} & Treats Bayesian uncertainty as a property of neural computation. \\
\addlinespace
\textit{``In short, hierarchical Bayesian approaches demand our accounts of cognition become deeper and better integrated.''} & Treats Bayesian modeling as a methodological framework constraining cognitive theorizing. \\
\addlinespace
\textit{``These considerations, which concern all Bayesian models of psychophysical data, highlight the gap between normative descriptions and their biological implementation.''} & Identifies a gap between formal Bayesian description and biological mechanism. \\
\bottomrule
\end{tabular}
\caption{Example assumption-bearing quotes. Each makes an implicit or explicit claim about what Bayesian modeling accomplishes for cognitive explanation, making it codable on the Realism/Instrumentalism scales.}
\label{tab:abq}
\end{table}

\subsection{Deployment Corpus}
\label{sec:dep}

The deployment corpus was assembled in three stages.
\textbf{(i)~Article selection.} We searched the APA PsycInfo database with keywords relevant to Bayesian modeling in cognitive science and collected 225 candidate articles spanning 1983--2025.
\textbf{(ii)~Quote extraction.} For each article, three frontier LLMs (GPT-5.2 \citep{openai2025gpt52}, Claude Opus 4.6 \citep{anthropic2026opus46}, Gemini 3 Pro Preview \citep{google2025gemini3pro}, the heavier-tier variants used only for corpus construction) received the 251 expert-extracted quotes and 15 articles as in-context examples, and were prompted to extract assumption-bearing quotes from the article's full text.
The same three models voted on each article's topical relevance; articles receiving 0/3 or 1/3 votes were dropped, leaving 210 articles.
\textbf{(iii)~Fidelity check.} Extracted quotes were verified against the source PDF by a string-matching script; quotes detected as hallucinated or paraphrased were dropped.
A manual audit of a uniform 70-quote sample found zero fidelity errors.
The final deployment corpus contains 6{,}858 verified assumption-bearing quotes from 210 articles.
The final-annotation stage (§\ref{sec:method}) uses the lighter-tier variants within the same three providers (GPT-5.1 \citep{openai2025gpt51}, Claude Sonnet 4.6 \citep{anthropic2026sonnet46}, and Gemini 3 Pro Preview, the same as in construction) so that the headline reliability reflects the cheaper, more widely deployable model within each provider; the heavier variants in the construction stage were chosen because corpus construction is a one-time pass rather than an iterated optimization loop.

\section{Method}
\label{sec:method}

Figure~\ref{fig:workflow} summarizes the study workflow: a theory-driven codebook and expert-coded reference annotations feed an autoresearch prompt-optimization loop (detailed in Figure~\ref{fig:autoresearch}), yielding a final shared zero-shot prompt that is deployed to three frontier LLMs for annotation and multi-level reliability analysis.

\begin{figure*}[t]
\centering
\includegraphics[width=\textwidth]{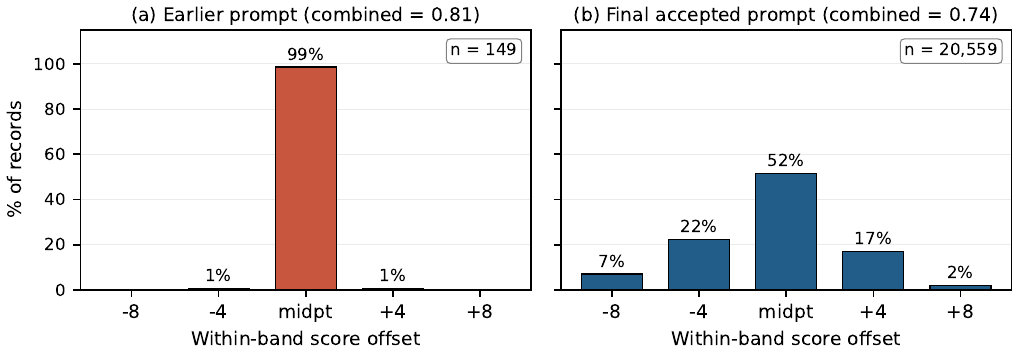}
\caption{Within-band score-offset distributions pooled across the three raters, restricted to the codebook-allowed positions $\{-8, -4, 0, +4, +8\}$. \textbf{(a)} An earlier prompt (combined~$=$~0.81) assigned 99\% of records to band midpoints. \textbf{(b)} The final accepted prompt uses all five offset positions, demonstrating real within-band scale use.}
\label{fig:midpoint}
\end{figure*}

\subsection{Prompt Development}

\begin{figure*}[t]
\centering
\begin{tikzpicture}[
  node distance=0.32cm and 0.40cm,
  font=\scriptsize,
  every node/.style={align=center},
  procstep/.style={rectangle, rounded corners=3pt, draw=black!70, fill=cyan!8,
                   minimum width=2.7cm, minimum height=0.75cm, inner sep=3pt},
  gate/.style={rectangle, rounded corners=3pt, draw=black!70, fill=yellow!15,
               minimum width=4.6cm, minimum height=1.55cm, inner sep=4pt,
               text width=4.45cm, align=left},
  phase/.style={rectangle, rounded corners=3pt, draw=black!70, fill=green!12,
                minimum width=2.6cm, minimum height=0.95cm, inner sep=3pt},
  expertbox/.style={rectangle, rounded corners=3pt, draw=black!70, line width=0.6pt,
                    fill=orange!18, minimum width=2.6cm, minimum height=1.3cm,
                    inner sep=3pt, text width=2.55cm, align=center},
  decision/.style={diamond, aspect=2.2, draw=black!70, fill=red!12,
                   minimum width=1.6cm, inner sep=1pt},
  exitbox/.style={rectangle, rounded corners=2pt, draw=black!50, dashed,
                  fill=gray!8, inner sep=4pt, text width=5cm, align=center},
  versionbox/.style={rectangle, rounded corners=3pt, draw=black!70, fill=gray!12,
                     minimum width=2.6cm, minimum height=0.95cm, inner sep=3pt},
  arr/.style={-Latex, thick, draw=black!70},
  larr/.style={-Latex, thick, draw=black!70, dashed},
]

\node[font=\small\bfseries] (A) {(a) Autoresearch loop within one version};
\node[procstep, below=0.25cm of A]                           (best)    {Best current\\prompt $P_n$};
\node[procstep, below=of best]                               (propose) {Agent proposes\\modification $P_{n+1}$};
\node[procstep, below=of propose]                            (run)     {Run $P_{n+1}$ on\\tuning split ($n{=}151$)};
\node[gate, below=of run]                                    (gates)   {\textbf{Validity gates:}\\[1pt]
  $\bullet$ $\Delta$\,combined\_metric $> 0$\\
  $\bullet$ off\_midpoint $\geq$ prior $- 0.02$\\
  $\bullet$ residual\_ICC $\geq$ max(prior $- 0.02$, $-0.05$)\\
  $\bullet$ gate\_persistent $\leq$ prior $+ 1$};
\node[decision, below=of gates]                              (dec)     {pass?};

\draw[arr] (best)    -- (propose);
\draw[arr] (propose) -- (run);
\draw[arr] (run)     -- (gates);
\draw[arr] (gates)   -- (dec);

\draw[arr]  (dec.east) -- ++(2.2,0) node[midway, above, font=\tiny] {accept ($P_{n+1}{\to}P_n$)}
                       |- (best.east);
\draw[larr] (dec.west) -- ++(-2.3,0) node[midway, above, font=\tiny] {reject (revert)}
                       |- (propose.west);

\node[exitbox, below=0.3cm of dec] (exit) {\textbf{Exit Phase 1 when:}\\soft plateau (3 consecutive rejects), hard plateau (5 iters w/o $\Delta$\,combined), or budget cap};
\draw[arr, dotted] (dec.south) -- (exit.north);

\node[font=\small\bfseries, right=2.0cm of A] (B) {(b) Phase progression and version transition};

\node[phase, below=0.25cm of B]             (p1)    {Phase 1\\ tuning iter loop\\ \scriptsize(panel a)};
\node[phase, below=of p1]                   (p2)    {Phase 2\\ dev validation\\ \scriptsize($n{=}50$, 3 seeds)};
\node[phase, below=of p2]                   (p3)    {Phase 3\\ held-out test\\ \scriptsize($n{=}50$, 2--3 seeds)};
\node[expertbox, below=of p3]               (audit) {\textbf{Expert audit}\\ \scriptsize inspect sample\\ \scriptsize classifications, audit\\ \scriptsize prompt, revise gates};
\node[versionbox, below=of audit]           (vnext) {Version $N{+}1$\\ \scriptsize(new chat session,\\baselined on $N$'s\\accepted prompt)};

\draw[arr] (p1)    -- node[right, font=\tiny, xshift=2pt] {best $P_n$} (p2);
\draw[arr] (p2)    -- node[right, font=\tiny, xshift=2pt] {generalization OK} (p3);
\draw[arr] (p3)    -- node[right, font=\tiny, xshift=2pt] {final report} (audit);
\draw[arr] (audit) -- node[right, font=\tiny, xshift=2pt] {revised instructions} (vnext);

\draw[larr] (p2.west) .. controls +(-1.7,0) and +(-1.7,0) ..
            node[midway, left, font=\tiny, xshift=-2pt] {dev gap too large}
            (p1.west);

\draw[larr] (vnext.east) .. controls +(1.4,1.5) and +(1.4,-1.5) ..
            node[midway, right, font=\tiny, xshift=2pt] {new chat} (p1.east);

\end{tikzpicture}
\caption{Autoresearch optimization process. \textbf{(a)} Within-version loop: a prompt edit is accepted only if it improves the combined metric \emph{and} passes all four validity gates. \textbf{(b)} Phase progression: the Phase-1 prompt is validated on dev and held-out test, with an expert audit between versions revising the agent's gates before the next version begins.}
\label{fig:autoresearch}
\end{figure*}
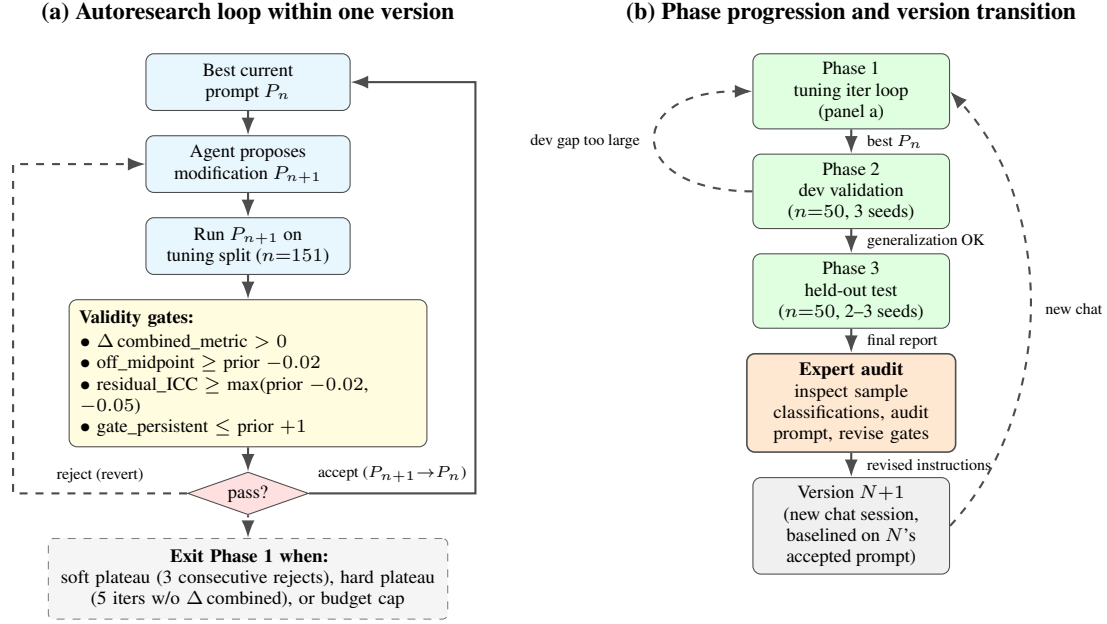

Prompt development adapted the autoresearch agent paradigm \citep{karpathy2026autoresearch}: an agent proposes modifications, runs experiments, and accepts only successful changes (Figure~\ref{fig:autoresearch}a). While automated prompt optimization itself is well established \citep[e.g.,][]{pryzant2023,khattab2024}, our use of an agent-driven loop gated by both a reliability metric and diagnostic checks follows the autoresearch design. We implemented this in Claude Code (Opus 4.7) \citep{anthropic2026opus47} as the optimization agent (not one of the final annotation raters), and ran more than fifty prompt-optimization attempts across twelve versioned rounds. Within a version, prompt changes were made only against the tuning split; the selected prompt was then evaluated on development (3 seeds) and held-out test (2--3 seeds) splits (Figure~\ref{fig:autoresearch}b; full per-version trajectory in Appendix~\ref{app:trajectory}). These repeated rounds constitute internal validation under low-label constraints, not a fully external benchmark.

We intentionally optimized one shared annotation prompt rather than separate prompts for each model, treating the prompt as a common coding protocol so agreement reflects convergence under shared instructions rather than model-specific tuning. The final prompt is zero-shot and codebook-guided; a worked-example variant was tested during development but rejected because it failed the acceptance criteria and increased diagnostic violations.

\subsection{Optimization Metric and Diagnostics}
\label{sec:metric}

The autoresearch loop requires a scalar optimization target, but qualitative coding has no objective validation score. We constructed one from our two primary reliability metrics, ICC(2,1) on the continuous Realism/Instrumentalism scores and Krippendorff's $\alpha$ on the ordinal band labels, combined via the harmonic mean to penalise edits that improve one at the expense of the other. Across 21 within-version edit deltas (Appendix~\ref{app:trajectory}), ICC and $\alpha$ moved together on 76\% of edits and traded off on 24\% (Pearson $r$ between $\Delta\mathrm{ICC}$ and $\Delta\alpha$ = 0.79), so the two metrics are correlated enough to combine while the trade-off cases justify the harmonic-mean penalty. Multi-objective enforcement is handled separately by the diagnostic gates below; Fleiss' $\kappa$ on band labels and within-band residual ICC (ICC(2,1) on within-band score offsets $r = s - m_b$, computed on quotes where rater and reference agree on band $b$ and averaged over scales then raters) were also logged as side-effect checks.

The checks were failure-driven rather than preregistered. In an earlier version, the held-out combined score reached 0.8068 but models showed near-universal midpoint behavior, agreeing by assigning most quotes to band midpoints rather than using the within-band score (Figure~\ref{fig:midpoint}). This shortcut motivated the development of diagnostic gates below.

\paragraph{Diagnostic gates.}
A candidate prompt edit was accepted only if it passed all four gates, each evaluated against the current best prompt's value (\emph{prior}) on the same tuning split:
\begin{itemize}\setlength{\itemsep}{1pt}\setlength{\topsep}{2pt}\setlength{\parsep}{0pt}
  \item \textbf{Combined metric}: $\Delta\,\mathrm{combined} > 0$ --- the candidate must improve the harmonic-mean score.
  \item \textbf{Off-midpoint}: $\mathrm{off\_midpoint\_pct} \geq \mathrm{prior} - 0.02$ --- the fraction of continuous scores not at the band midpoint may drop at most 2 points; explicit guard against the midpoint-collapse pattern in Figure~\ref{fig:midpoint}.
  \item \textbf{Within-band residual ICC}: $\mathrm{residual\_ICC} \geq \max(\mathrm{prior} - 0.02,\ -0.05)$ --- within-band continuous agreement may drop at most 2 points from the prior, with an absolute floor at $-0.05$.
  \item \textbf{Persistent coherence failures}: $\mathrm{gate\_persistent\_fails} \leq \mathrm{prior} + 1$ --- records failing the structural coherence checks (schema, score--band consistency, allowed tags per band) \emph{after} auto-correction may increase by at most one across the 151-quote tuning split.
\end{itemize}
The gates are one-sided floors that prevent metric-validity regressions without constraining legitimate improvements; other coding tasks would require different diagnostics.

A second class of diagnostic concerns construct validity. Earlier prompt versions leaked the expected distributional answer back into the model: the prompt specified Weak Realist as the default band, and it referenced approximate band frequencies that mirrored the expected corpus distribution. Both biases inflate apparent reliability by making the prompt anticipate the answer, and bias corpus-level findings in the direction the prompt encodes. We removed both in a deliberate de-leaking step (the v12 baseline in Table~\ref{tab:ablation}, which scored $\sim$0.04 combined points below its un-de-leaked predecessor); a prompt encoding the expected distribution cannot test the construct, so all headline results are computed under the de-leaked prompt.

\paragraph{Expert oversight.}
The loop runs autonomously within a session, but the broader optimization is expert-led: between sessions the expert inspected sample classifications, audited the prompt for circular cues, and revised the agent's gates when systematic issues were detected (Figure~\ref{fig:autoresearch}b). Both task-specific gates reported here (midpoint-collapse, de-leaking) were expert-discovered, because each problem inflated rather than degraded the headline metric.

\subsection{Raters and Deployment}

The final deployment used three independent LLM raters: GPT-5.1, Claude Sonnet 4.6, and Gemini 3 Pro Preview.
Each model returned structured JSON with band labels, continuous scores, language-criteria tags, distinguishing-criteria citations, and a one-sentence justification.
A coherence gate checked schema validity, score--band consistency, label-slot alignment, allowed tags per band, and bullet-verbatim citation format; records that failed gate checks after auto-correction were flagged for review but not dropped from the analysis.
We computed ICC(2,1), Krippendorff's $\alpha$ (ordinal), Fleiss' $\kappa$, pairwise band agreement, and per-rater article-level random-effects models.

\paragraph{Reproducibility.}
The three LLMs were accessed via their providers' commercial batch APIs (OpenAI, Anthropic, Google) under each provider's standard terms of service for research use; parameter counts are not published by the providers, and no local GPU compute was used.
Total API usage across all stages (autoresearch loop, corpus construction, and final deployment) was approximately 150 million tokens, run predominantly in batch mode.
Statistical analyses were implemented in Python using \texttt{statsmodels} (mixed-effects models), \texttt{scipy} (inference), and \texttt{NumPy/pandas} (data handling).
Code, codebook, and reference annotations will be released in the project repository.

\section{Results}

\subsection{Prompt Search and Held-Out Validation}

The final prompt search began from a human-developed baseline that already incorporated earlier codebook and prompt revisions, including a deliberate de-leaking step that removed band-frequency claims and default-band cues to avoid circular validation (Section~\ref{sec:metric}).
Table~\ref{tab:ablation} summarizes the final ablation: accepted changes were retained; rejected changes were reverted before the next attempt.

The accepted prompt reached a development-set combined score of 0.812 (sd = 0.013; tuning--dev gap = 0.021, mean of three seeds).
On the held-out test split (mean of two seeds), it reached 0.763 (sd = 0.014), with ICC = 0.79, $\alpha$ = 0.74, and Fleiss' $\kappa$ = 0.57.
The dev--test gap (0.049) is consistent with small-sample variance on a 50-quote partition; all diagnostics were satisfied.

\begin{table}[t]
\centering
\small
\setlength{\tabcolsep}{6pt}
\renewcommand{\arraystretch}{1.12}
\begin{tabular}{@{}rllr@{}}
\toprule
\textbf{Iter.} & \textbf{Variant} & \textbf{Decision} & \textbf{Score} \\
\midrule
1 & baseline (de-leaked) & ---             & 0.768 \\
\textbf{2} & \textbf{wr\_rs\_boundary} & \textbf{ACCEPT} & \textbf{0.791} \\
3 & asif\_rs              & reject          & 0.789 \\
4 & mr\_entry             & reject          & 0.775 \\
5 & worked\_examples      & reject; plateau & 0.774 \\
\bottomrule
\end{tabular}
\caption{Final prompt-search ablation (v12) on the tuning split. The single accepted change (bold) clarified the Weak-Realist vs.\ Realism-Sceptical boundary by routing ``describing/applying the model'' cases to Weak Realist; three subsequent revisions failed to clear the diagnostic gates and were reverted.}
\label{tab:ablation}
\end{table}

\subsection{Inter-Model Agreement on the Deployment Corpus}

On the 6{,}858 quotes annotated by all three LLMs, multi-rater agreement was substantial: ICC = 0.799 [95\% CI: 0.780, 0.818], $\alpha$ = 0.762 [0.736, 0.787], Fleiss' $\kappa$ = 0.579 [0.559, 0.597], and combined score = 0.780.
Confidence intervals are article-clustered bootstrap estimates (200 resamples), reflecting that quotes within an article are not independent.
Pairwise exact agreement was 78.4\%, and 88.4\% of pairwise comparisons were either exact or within one adjacent band, indicating that most disagreement is local rather than categorical.
Pairwise rates were similar across model pairs (Table~\ref{tab:pairwise}), suggesting broad convergence under the shared prompt.

Reliability is band-dependent: on the dominant Weak Realist and Moderately Realist categories ($n=5{,}911$), continuous-score ICC = 0.83 and $\alpha$ = 0.78; on the conceptual middle (Realism-Sceptical, Agnostic, Leaning Realist; $n=859$) the same statistics drop to 0.14 and 0.16; on the rare extreme bands ($n=88$) raters agree on the continuous score (ICC = 0.94) but split on the band boundary ($\alpha$ = 0.30).
The middle-band weakness reflects substantive interpretive difficulty: hedged or ambiguous passages where the codebook's promotion triggers are designed to fire but the cue itself is borderline (§\ref{sec:disagree}); we read this as the kind of disagreement human coders would also produce on the same material, although our single-expert reference annotations do not let us verify that claim (§\ref{sec:limits}).
Collapsing the seven bands to a three-tier scheme (extremes / middle / dominant), all three raters land in the same tier 75.2\% of the time and at least two of three do 99.8\% of the time (PABAK = 0.75 on the 3-tier collapse).
A permutation test that shuffles band labels within each rater (200 reps) places the null ICC at $0.000 \pm 0.014$ and the null $\alpha$ at $0.000 \pm 0.015$; the observed values sit more than 50 standard deviations above the null.
Appendix~\ref{app:reliability} reports the full per-band confusion, the 3-tier confusion matrix, and the permutation distribution.

\begin{table}[t]
\centering
\small
\setlength{\tabcolsep}{8pt}
\renewcommand{\arraystretch}{1.12}
\begin{tabular}{@{}lcc@{}}
\toprule
\textbf{Pair} & \textbf{Exact (\%)} & \textbf{Within-1 (\%)} \\
\midrule
GPT-5.1 vs.\ Claude   & 78.1 & 88.2 \\
GPT-5.1 vs.\ Gemini   & 78.0 & 87.5 \\
Claude  vs.\ Gemini   & 79.0 & 89.3 \\
\bottomrule
\end{tabular}
\caption{Pairwise band-agreement percentages. \emph{Within-1} counts exact or adjacent-band agreement on the 7-band Realism scale.}
\label{tab:pairwise}
\end{table}

\subsection{Article-Level Reliability and Model Differences}
\label{sec:article}

\begin{figure*}[t]
\centering
\includegraphics[width=\textwidth]{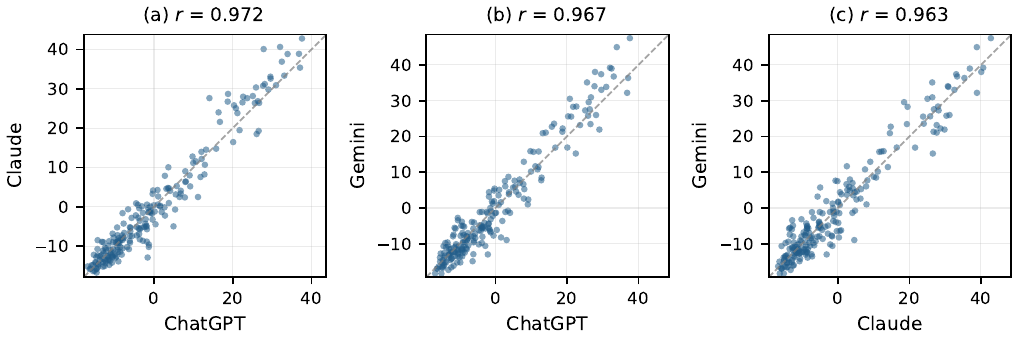}
\caption{Article-level Pearson correlations between rater pairs of per-article random intercepts. Each point is one of 210 articles; dashed line is identity. Article-level agreement ($r$ = 0.96--0.97) is substantially higher than quote-level ($\alpha$ = 0.76): the three LLMs converge on ranking articles even where they disagree on individual quotes.}
\label{fig:blup}
\end{figure*}

At the quote level, the three models were not interchangeable: relative to GPT-5.1, Claude scored Realism lower by 4.56 points and Gemini by 4.14. These are systematic calibration shifts, not disagreement about which articles are more realist. The shifts have band-specific signatures: Claude over-applies Realism-Sceptical (1.8$\times$ corpus mean), Gemini over-applies Strong Realism (1.8$\times$) and Leaning Realist (1.4$\times$), and GPT-5.1 over-uses Agnostic (2.0$\times$), characterising the per-model ``temperaments'' that emerge under the shared prompt.

To separate calibration from article-level ordering, we fit per-rater mixed-effects models on the Realism score with article as random intercept and extracted each rater's per-article random intercepts. Cross-rater Pearson correlations of these intercept estimates ranged from 0.963 to 0.972 (Figure~\ref{fig:blup}); Spearman $\rho$ from 0.908 to 0.948; Kendall's $\tau$ from 0.752 to 0.814 with bootstrap 95\% CIs above 0.70 for all pairs. An independent crossed-effects model (quotes nested in articles, raters crossed) attributes 32.0\% of total score variance to article and only 0.9\% to rater, confirming that the per-rater separation is recovering a genuine article-level signal. The divergence between the 0.76 quote-level $\alpha$ and the 0.96+ article-level rank stability is itself a finding: agreement at the grain that matches the downstream inference (the article, for corpus-mapping uses) recovers what a single-grain summary obscures.

The corpus was overall weakly realist with an instrumentalist tilt: Weak Realist covered 67.9\% of quotes and was the modal band for 81.0\% of articles; Moderately Realist was second (18.3\% of quotes, 17.6\% of articles). Yet articles were rarely single-stance: only 3 of 210 (1.4\%) used a single band, while 125 (59.5\%) spanned four or more, so the field is weakly committed in central tendency but internally heterogeneous within papers.

The corpus also reveals a structural difference between subfields: per-article mean Realism scores are 8.8 points higher in low-level perception/motor articles than in high-level cognition articles (39.2 vs.\ 30.4 on the 0--100 Realism scale; $n=115$ vs.\ $n=95$; Welch's $t=4.38$, $p<.001$; Cohen's $d=0.60$ on per-article means, $d=0.30$ on the underlying quote-level distribution). An LMM with rater fixed effect and article random intercept gives an equivalent $+8.81$-point coefficient ($95\%$ CI $[4.76, 12.86]$, $p<.001$); the effect is rater-independent (leave-one-rater-out coefficients in $[+8.5, +9.0]$, all $p<.001$; Appendix~\ref{app:subfield}). This quantifies a long-held qualitative intuition that low-level cognition researchers commit more strongly to mechanistic interpretation of Bayesian models than high-level cognition researchers, and illustrates how reliable corpus-scale annotation can supply empirical evidence for previously hand-asserted claims.

\begin{table*}[t]
\centering
\small
\setlength{\tabcolsep}{6pt}
\renewcommand{\arraystretch}{1.15}
\begin{tabular}{@{}p{0.20\textwidth}p{0.48\textwidth}ccc@{}}
\toprule
\textbf{Pattern} & \textbf{Quote} & \textbf{GPT-5.1} & \textbf{Claude} & \textbf{Gemini} \\
\midrule
Future-work mechanism & \textit{``Future studies are necessary to examine whether and how the two counteracting systems are implemented in the brain\ldots''} & MR (77) & LR (60) & Agn (50) \\
\addlinespace
Formal object as mental state & \textit{``\ldots neurons `must represent probability distributions' and `must be able' to implement Bayesian inference.''} & MR (85) & RS (36) & MR (81) \\
\addlinespace
Hedged interpretation & \textit{``The success of this Bayesian formalization may be interpreted as an optimal adaptation of our perceptual system\ldots''} & WR (23) & LR (64) & WR (15) \\
\bottomrule
\end{tabular}
\caption{Three high-disagreement quotes. WR = Weak Realist, RS = Realism-Sceptical, Agn = Agnostic, LR = Leaning Realist, MR = Moderately Realist; parentheticals are within-band scores.}
\label{tab:disagree}
\end{table*}

\subsection{Where the Raters Disagree}
\label{sec:disagree}

The 17.4\% of quotes where raters span more than one band typically share recognizable structural features.
Table~\ref{tab:disagree} shows three illustrative cases, each representing a distinct disagreement pattern.

These cases share a common structure: each contains a cue that the codebook's promotion triggers are designed to disambiguate, but where the cue itself is borderline.
Pattern 1 splits between MECH-subject promotion and Agnostic neutrality.
Pattern 2 splits between an evaluative-stance reading and a direct MECH+STRONG reading.
Pattern 3 splits on whether the hedged framing counts as AS-IF (Weak Realist) or as a directional implementation hint (Leaning Realist).
Disagreement is therefore structured around codebook-anticipated boundaries rather than random.

\section{Discussion}

The paper demonstrates that frontier LLMs can support theoretically demanding qualitative coding when the construct is operationalized through an auditable codebook, validated against expert reference annotations, and gated by task-specific diagnostics that catch failure modes the headline metric cannot. 

The diagnostic-aware autoresearch loop delivered a deployable prompt at a fraction of manual cost, but the two gates that prevented the combined metric from being gamed (off-midpoint, anti-circularity de-leaking) were both expert-discovered between sessions, which is why we frame the framework as a tool that extends rather than replaces expert judgment.

The shared-prompt protocol across three providers produced substantial convergence under a common codebook while preserving distinct calibration ``temperaments'' (Claude tilts toward Realism-Sceptical, Gemini toward Strong Realism, GPT-5.1 toward Agnostic). This convergence is most decisive at the article level, where the quote-level $\alpha$ of 0.76 rises to article-level rank correlations of 0.96--0.97 (Kendall's $\tau$ = 0.75--0.81 on the same random intercepts), with an independent crossed-effects model attributing 32\% of total variance to article and only 0.9\% to rater; reporting agreement at the grain that matches the downstream inference recovers what a single-grain summary obscures. 

The codebook design pattern that supports this convergence (complementary continuous scales constrained to sum to 100, ordinal bands with within-band scoring positions, and explicit promotion triggers making boundary decisions auditable) addresses several known pitfalls of operationalizing philosophically complex constructs in one integrated structure; it is a synthesis assembled for one demanding case rather than a universal template.

Beyond the methodology, the substantive picture of Bayesian cognitive science illustrates what reliable corpus-scale annotation can recover: a field weakly committed in central tendency (mostly Weak Realist), internally heterogeneous within papers (67.8\% of Realism-score variance within articles; 59.5\% of articles span four or more bands), and structurally divided between subfields (low-level perception and motor articles score 8.8 Realism points higher than high-level cognition articles, $p<.001$, $d=0.60$ on per-article means). Recovering this long-held but previously hand-asserted intuition with quantitative evidence is itself an indication that the method served the question. Future work will adapt the codebook architecture, the diagnostic-gate design, and the shared-prompt multi-rater protocol to other theoretically demanding qualitative coding tasks, ideally with multi-expert reference annotations to support stronger validity claims.

\section{Limitations}
\label{sec:limits}

The deployment corpus covers a single subdomain (Bayesian cognitive science, 210 articles, 6{,}858 quotes spanning 1983--2025), and the codebook was developed for this domain. The 210-article corpus is a snapshot of the field rather than an exhaustive sample (many additional Bayesian cognitive science articles exist outside our APA PsycInfo keyword search), so corpus-level findings describe this snapshot rather than the field as a whole.

Within each provider, we used a mid-tier rather than top-tier model variant primarily for budget reasons (e.g., Claude Sonnet rather than Opus) and did not systematically compare reliability across model sizes within a provider; whether larger variants would shift the headline reliability is an open question.

Expert-coded reference annotations are available for 251 quotes from one domain expert because additional domain experts trained on this codebook were not available within the project's resource and timeline constraints. We use the 251 quotes as an internal validation set rather than a multi-expert adjudicated benchmark, and we cannot estimate human--human reliability from this single-coder pool. As a complementary content check beyond the metric numbers, the same expert manually audited a 100-classification sample drawn from the deployment corpus for both formal coherence (schema, score--band consistency) and content fidelity (band assignment justified by the quote); the audit found no systematic departures from the codebook.

The deployment corpus is itself shaped by LLM-assisted quote extraction and topical-relevance voting (§\ref{sec:dep}); we did not run a rule-based extraction sensitivity check, a comparison against fine-tuned classifiers, or chain-of-thought / chain-of-stance prompting variants beyond the rejected worked-examples diagnostic, and we did not run a pinned drift audit across API updates.

High inter-model agreement may partly reflect shared training-data biases rather than fully independent evidence; the three models are not three independent expert coders. We mitigate this by requiring the codebook to drive the classification, reporting model-specific calibration shifts, and reporting agreement at multiple grain sizes, but the concern cannot be fully resolved without external human comparison. The methodology gate flagged but retained 1{,}292 records (6.3\% of classifications) on a flag-don't-drop principle; metric sensitivity to this choice was small ($<$0.01 in headline reliability). The codebook and diagnostic-gate design could be adapted to detect stance on contested non-science topics (e.g., political ideology), where automated stance classification at scale raises governance concerns that the present technical scope does not address.

\paragraph{Use of AI assistants.}
The autoresearch prompt-optimization agent (§\ref{sec:method}) was implemented in Claude Code using Claude Opus 4.7; this is part of the paper's methodological contribution rather than an undisclosed authoring tool. We additionally used Claude (Opus 4.7) as a writing assistant for drafting and revising the manuscript, and as a coding assistant for implementing the Python analysis scripts (\texttt{statsmodels}, \texttt{scipy}, \texttt{NumPy/pandas}). The statistical computations themselves are deterministic and performed by these standard libraries; all numerical results were verified by the author against the raw data, and all scientific claims, analyses, and interpretations are the author's own.

\bibliography{custom}

\clearpage
\appendix

\section{Codebook and Annotation Prompts}
\label{app:codebook}

This appendix shows the codebook architecture (Sec.~\ref{app:codebook-arch}) with the full Realism scale reproduced in Figure~\ref{fig:codebook} (Sec.~\ref{app:codebook-full}), the per-quote user-prompt template (Sec.~\ref{app:userprompt}), and the procedural component of the system prompt (Sec.~\ref{app:taskmd}).
The codebook component of the system prompt comprises seven Realism bands and seven complementary Instrumentalism bands; Figure~\ref{fig:codebook} shows the Realism scale, with the Instrumentalism scale mirroring the same structure (full codebook in the project repository).
The 251 reference annotations, the full deployment classifications, and the analysis code will be released at the project repository upon acceptance.

\subsection{Codebook Architecture}
\label{app:codebook-arch}

The codebook is organized as two complementary ordinal scales, Realism and Instrumentalism.
Each scale has seven bands with explicit numeric ranges:

\begin{center}
\small
\begin{tabular}{@{}lcc@{}}
\toprule
\textbf{Realism band} & \textbf{Range} & \textbf{Midpoint} \\
\midrule
Anti-Realist         &   0--10  &   5 \\
Weak Realist         &  11--27  &  19 \\
Realism-Sceptical    &  28--44  &  36 \\
Agnostic             &  45--55  &  50 \\
Leaning Realist      &  56--72  &  64 \\
Moderately Realist   &  73--89  &  81 \\
Strong Realism       &  90--100 &  95 \\
\bottomrule
\end{tabular}
\end{center}

Each Realism band has a complementary Instrumentalism band whose midpoint sums to 100 (e.g., Moderately Realist (81) pairs with Weak Instrumentalist (19)).
Within each band, raters choose one of five score positions: midpoint, midpoint $\pm 4$, or midpoint $\pm 8$.

Each band is specified along four classification axes:
\textbf{Subject} (canonical tags: \texttt{MODEL}, \texttt{AGENT}, \texttt{MECH}, \texttt{MIXED});
\textbf{Claim Target} (\texttt{FUNC}, \texttt{BEH}, \texttt{MAP}, \texttt{MECH});
\textbf{Modality} (\texttt{NEUTRAL}, \texttt{HEDGED}, \texttt{STRONG}, \texttt{AS-IF});
and \textbf{Evidence Link} (\texttt{NONE}, \texttt{INFERRED}, \texttt{CONC}, \texttt{DIRECT}).
Bands provide separate Upper-Band and Lower-Band tag specifications for each axis, used during cue-audit scoring (Section~\ref{sec:method}).

Five Realism bands are reached by \emph{promotion triggers}, explicit rule patterns that fire when a recognizable cue is present.
For example, Realism-Sceptical requires one of three triggers: a mapping claim without mechanism, a Bayesian account framed as an alternative to a posited mechanism, or a Bayesian formal object directly equated with an agent's mental state.
Promotion triggers are checked first; if none fires, the band is read off a band matrix using the quote's Subject/Target/Modality/Evidence-Link profile.

\subsection{Codebook: Realism Scale}
\label{app:codebook-full}

\begin{figure*}[h!]
\centering
\includegraphics[width=\textwidth]{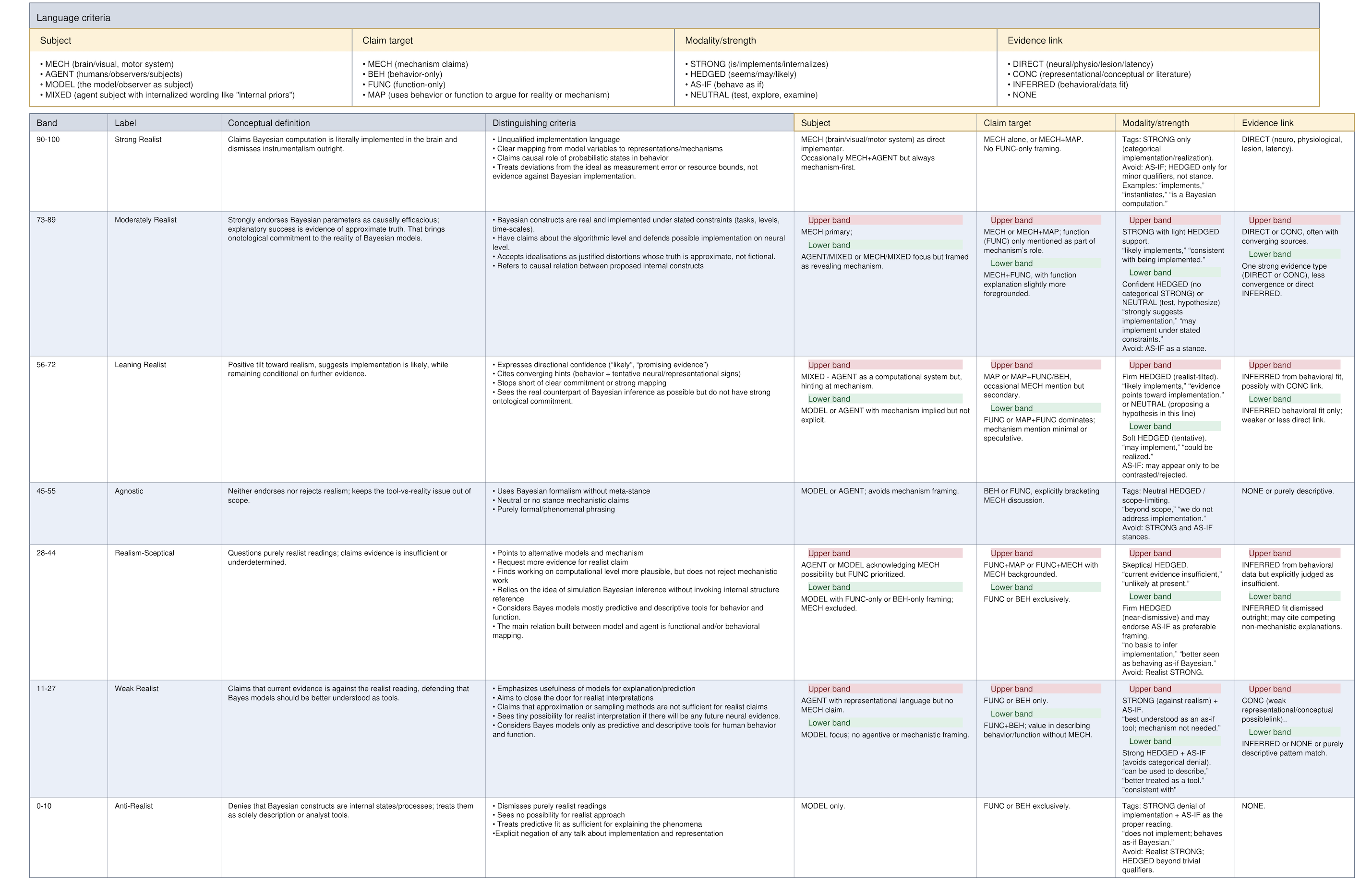}
\caption{The seven-band Realism scale of the codebook. Columns: the four classification axes (Subject, Claim Target, Modality, Evidence Link) plus Distinguishing Criteria. Rows: the seven Realism bands. Each cell gives Upper-Band and Lower-Band specifications used during cue-audit scoring. The complementary Instrumentalism scale (not shown) mirrors this structure; the two scores per quote sum to 100. Full codebook in the project repository.}
\label{fig:codebook}
\end{figure*}

Figure~\ref{fig:codebook} reproduces the Realism scale of the codebook, defining all seven Realism bands.
Each band specifies a numeric scoring range, a Conceptual Definition, Distinguishing Criteria bullets, and four Language Criteria specifications (Subject, Claim Target, Modality, Evidence Link) with Upper-Band and Lower-Band specifications used during the cue-audit scoring step (Section~\ref{sec:method}).
The complementary Instrumentalism scale mirrors this structure: each Realism band is paired with an Instrumentalism band whose midpoint sums to 100, with the same four-axis Language Criteria specification per band.
The full codebook (both scales) is available in the project repository.

\subsection{Per-Quote User Prompt Template}
\label{app:userprompt}

For each quote, the user-prompt wrapper around the quote text was as follows.
The OpenAI (GPT-5.1) and Anthropic (Claude Sonnet 4.6) rater paths used the standard template; the Gemini (Gemini 3 Pro Preview) path used a variant that asks the model to echo the \texttt{quote\_id} in its JSON output, because Gemini's batch API stopped echoing request-metadata in responses during the deployment window.

\paragraph{Standard template (OpenAI, Anthropic):}
\begin{Verbatim}[fontsize=\footnotesize, frame=single, framesep=4pt, breaklines=true]
Quote ID: {quote_id}

Classify the following quote:

"{quote}"
\end{Verbatim}

\paragraph{Gemini template:}
\begin{Verbatim}[fontsize=\footnotesize, frame=single, framesep=4pt, breaklines=true]
Quote ID: {quote_id}

REQUIRED: your JSON response MUST include the field
"quote_id": {quote_id} (any position in the JSON object is
fine, but it must be present).

Classify the following quote:

"{quote}"

Reminder before you respond: include "quote_id": {quote_id}
in your JSON output, this is how we link the response back
to the right quote.
\end{Verbatim}

\subsection{Procedural Component of System Prompt}
\label{app:taskmd}

The system prompt sent to each LLM was the concatenation of (i) the procedural component shown below (the contents of \texttt{task.md}) and (ii) the codebook component (the contents of \texttt{codebook.md}; Realism scale shown in Figure~\ref{fig:codebook}).

\begin{Verbatim}[fontsize=\scriptsize, frame=single, framesep=4pt, breaklines=true]
## Task

You classify quotes about Bayesian cognitive science along two scales:
Realism and Instrumentalism. For each scale, pick one of seven ordinal
bands and assign a continuous score within the band's range. Use only the
codebook's Conceptual Definitions, Distinguishing Criteria, and Language
Criteria. Do not introduce categories the codebook does not define.

The two scores must sum to 100. Each score must lie inside its band's
numeric range.

The codebook (codebook.md) is the authoritative spec for axis tags,
per-band Language Criteria, and Distinguishing Criteria. This file
(task.md) covers only the procedure: band selection, scoring, pairing,
and output format.

## Scales and bands

[7-row table of Realism bands paired with complementary Instrumentalism
 bands; reproduced in Appendix A.1.]

Pairing. The two scales are complementary: pick the Realism band from
the quote's cues, then take its paired Instrumentalism band from the
table above.

## The four classification axes

S = Subject, T = Claim Target, M = Modality, L = Evidence Link, and
their canonical tags are defined in codebook.md (Language Criteria).

## Band-selection logic

### Band triggers (apply first)
[Six promotion triggers, one per band reached by trigger:
 Realism-Sceptical (Triggers A/B/C), Leaning Realist, Moderately Realist,
 Strong Realism, Anti-Realist, Agnostic.]

### Band matrix (apply after the promotion-trigger check)
[10-row matrix mapping Subject x Target x Modality patterns to candidate
 Realism bands.]

### Boundary-case rules
- Behavioral match is not a mechanistic claim.
- Climb to Leaning Realist or higher only when the quote says something
  about what is happening inside the brain/mind.
- Realism-Sceptical requires Trigger A, B, or C.
- Describing or applying the model is Weak Realist, not R-S.
- Moderately Realist requires MECH subject + a direct internal-state
  assertion.
- Anti-Realist requires an explicit disclaimer.

## Per-band scoring rubric

For the assigned band, mark each of the four axes (S/T/M/L) as
Lower-leaning, Upper-leaning, Neutral, or Conflicting using the band's
Language Criteria. Apply a count rule that maps {Upper count, Lower
count} to {midpoint, midpoint +/- 4, midpoint +/- 8}, capped at the
band's numeric range.

## Classification procedure

1. Pick the Realism band using triggers + band matrix + boundary rules.
2. Score the Realism band using the cue audit (count rule above).
3. Pick the Instrumentalism band as the complementary pair.
4. Consistency check: realism_score + instrumentalism_score == 100;
   nudges mirror across scales.

## Output discipline

The response must be a single JSON object with exactly eleven keys, in
fixed order, with no prose before { or after }:

{
    "input_quote": "",
    "predicted_realism_band_label": "Weak Realist",
    "predicted_instrumentalism_band_label": "Moderately Instrumentalist",
    "predicted_realism_d_c": "<verbatim Distinguishing Criteria bullet>",
    "predicted_instrumentalism_d_c": "<verbatim Distinguishing Criteria bullet>",
    "predicted_realism_lc_code": "S: <tag>\nT: <tag>\nM: <tag>\nL: <tag>",
    "predicted_instrumentalism_lc_code": "S: <tag>\nT: <tag>\nM: <tag>\nL: <tag>",
    "predicted_realism_score": 23,
    "predicted_instrumentalism_score": 77,
    "general_justification": "One short sentence tying the quote to the band.",
    "consistency_check": "TRUE"
}

Field rules (full constraints in the repository version of task.md):
- input_quote: empty string; the harness fills it in.
- band labels: exactly one of the seven labels for the relevant scale.
- d_c: a character-for-character copy of one bullet from the assigned
  band's Distinguishing Criteria list.
- lc_code: four lines, one canonical tag per axis with a brief cue.
- scores: must lie inside band range and equal {midpoint, +/-4, +/-8};
  sum to exactly 100.
- general_justification: one sentence <= 25 words.
- consistency_check: "TRUE" or "FALSE".

Total output should fit under 500 tokens.
\end{Verbatim}

\noindent The verbatim version of \texttt{task.md} (with full tables for the band-pairing scale, the promotion triggers, and the band matrix, and full field-rule text) is in the project repository.

\section{Reliability Robustness Checks}
\label{app:reliability}

This appendix reports additional diagnostics that evaluate the headline reliability in §5.2 against known confounds: band-frequency stratification, prevalence inflation, boundary-band disagreement structure, and chance baselines.
All numbers are computed on the full 6{,}858-quote deployment corpus annotated by all three raters.

\paragraph{Stratification by conceptual band group.}
The seven bands have a natural conceptual grouping on the underlying Realism dimension: the two dominant practical positions (Weak Realist, Moderately Realist), the three middle uncertainty bands (Realism-Sceptical, Agnostic, Leaning Realist), and the two polar extremes (Anti-Realist, Strong Realism).
Restricting to quotes whose fused (rater-median) band falls in each group:

\begin{center}
\small
\setlength{\tabcolsep}{6pt}
\renewcommand{\arraystretch}{1.1}
\begin{tabular}{@{}lrrrr@{}}
\toprule
\textbf{Group} & \textbf{n} & \textbf{ICC} & \textbf{$\alpha$} & \textbf{$\kappa$} \\
\midrule
Dominant (WR, MR)              & 5{,}911 & 0.833 & 0.780 & 0.634 \\
Middle (RS, Ag, LR)            &   859 & 0.143 & 0.158 & 0.121 \\
Extremes (AR, SR)              &    88 & 0.938 & 0.295 & 0.041 \\
\bottomrule
\end{tabular}
\end{center}

The dominant group's strong reliability and the middle group's weakness are both substantive findings: the codebook resolves typical practical positions cleanly but does not fully separate the three middle uncertainty bands.
The extreme group's split pattern (high score ICC, low band $\kappa$) reflects raters agreeing on the continuous extremity of the position while disagreeing on which side of the band boundary it sits, which is a continuous-vs-categorical artifact rather than a substantive disagreement and justifies our use of ICC on the continuous scores as the primary reliability metric.

\paragraph{Three-tier collapse.}
Collapsing the seven bands to three super-bands (extremes / middle / dominant), all three raters land in the same super-band on 75.2\% of quotes; at least two of three agree on 99.8\%.
On this 3-band collapse, Krippendorff's $\alpha$ = 0.346, Fleiss' $\kappa$ = 0.353, and PABAK = 0.751.
The gap between $\kappa = 0.35$ and PABAK = 0.75 reflects the heavy prevalence of ``dominant'' (86\% of quotes), which inflates the chance-agreement term in $\kappa$ but is corrected by PABAK.

The aggregate 3$\times$3 super-band confusion matrix (rater pair counts, summed across the three pairs) localises the disagreement to the middle-dominant boundary:

\begin{center}
\small
\setlength{\tabcolsep}{6pt}
\renewcommand{\arraystretch}{1.1}
\begin{tabular}{@{}lrrr@{}}
\toprule
            & \textbf{extr.} & \textbf{mid.} & \textbf{dom.} \\
\midrule
\textbf{extremes}  &   256 &    43 &   491 \\
\textbf{middle}    &    43 & 2{,}368 & 2{,}885 \\
\textbf{dominant}  &   491 & 2{,}885 & 31{,}686 \\
\bottomrule
\end{tabular}
\end{center}

The dominant cell shows where raters agree (31{,}686 pair-agreements on ``dominant'').
The largest off-diagonal cell is middle$\leftrightarrow$dominant (2{,}885 pair disagreements): when a quote is in the middle uncertainty zone for any rater, another rater frequently reads it as dominant practical position.

\paragraph{Per-band misclassification distance.}
Mean $|\Delta\mathrm{band}|$ per band (averaged across the three rater pairs) shows where on the 7-band scale disagreement concentrates:

\begin{center}
\small
\setlength{\tabcolsep}{6pt}
\renewcommand{\arraystretch}{1.05}
\begin{tabular}{@{}lrr@{}}
\toprule
\textbf{Band} & \textbf{n assignments} & \textbf{mean $|\Delta b|$} \\
\midrule
Anti-Realist                  &   112 & 1.00 \\
Weak Realist                  & 27{,}472 & 0.25 \\
Realism-Sceptical             &  1{,}730 & 0.93 \\
Agnostic                      &  1{,}048 & 1.50 \\
Leaning Realist               &  2{,}518 & 1.21 \\
Moderately Realist            &  7{,}590 & 0.59 \\
Strong Realism                &    678 & 0.82 \\
\bottomrule
\end{tabular}
\end{center}

Weak Realist is the most reliable band (mean band-distance 0.25); Agnostic and Leaning Realist are the least reliable (1.50 and 1.21), consistent with the stratified analysis above.

\paragraph{Permutation null.}
We permuted band labels within each rater 200 times and recomputed reliability metrics, holding marginal distributions fixed.
The null mean ICC was $+0.000$ with 95\% bounds $[-0.014, +0.013]$; null $\alpha$ was $-0.000$ with $[-0.015, +0.014]$; null $\kappa$ was $-0.002$ with $[-0.011, +0.006]$.
The observed values (ICC = 0.799, $\alpha$ = 0.762, $\kappa$ = 0.579) sit more than 50 standard deviations above the null distribution, ruling out chance.

\section{Subfield-Finding Robustness}
\label{app:subfield}

This appendix reports the alternative specifications used to verify the $+8.8$-point low-level vs.\ high-level Realism difference reported in §5.3.
All tests were run on the full 6{,}858-quote $\times$ 3-rater deployment data restricted to articles whose \texttt{High/Low} field is \texttt{Low-Level} (115 articles, 3{,}706 quote-rater rows per rater) or \texttt{High-Level} (95 articles, 3{,}152 rows per rater).

\paragraph{Linear mixed model.}
With rater as fixed effect and article as random intercept (\texttt{realism\_score} $\sim$ \texttt{high\_low} $+$ \texttt{rater} $+\,(1\,|\,\texttt{article\_id})$), the Low-Level vs.\ High-Level coefficient is $+8.81$ (SE = 2.07, 95\% CI $[4.76, 12.86]$, $p = 2.0 \times 10^{-5}$).
Rater fixed effects relative to GPT-5.1 are $-4.56$ (Claude, $p<10^{-22}$) and $-4.14$ (Gemini, $p<10^{-22}$); the article-level random-effect variance is 216.5.

\paragraph{Per-rater Welch's $t$.}
Run separately per rater on the quote-level scores:

\begin{center}
\small
\setlength{\tabcolsep}{6pt}
\renewcommand{\arraystretch}{1.05}
\begin{tabular}{@{}lrrrl@{}}
\toprule
\textbf{Rater} & \textbf{$\Delta M$} & \textbf{$t$} & \textbf{$d$} & \textbf{95\% CI} \\
\midrule
GPT-5.1 & $+7.95$ & 12.39 & 0.30 & $[6.69, 9.21]$ \\
Claude  & $+9.06$ & 14.97 & 0.36 & $[7.87, 10.24]$ \\
Gemini  & $+8.13$ & 12.06 & 0.29 & $[6.81, 9.45]$ \\
\bottomrule
\end{tabular}
\end{center}

All three $p$-values are $<10^{-32}$. The quote-level $d$ is the small-to-medium effect that absorbs within-article noise; the per-article $d$ reported in the main text (0.60) is the same difference standardised by the between-article SD.

\paragraph{Cluster bootstrap.}
We resampled articles with replacement (2{,}000 reps) and within each sampled article resampled its quote-rater rows with replacement.
The cluster-bootstrap mean of $\Delta M$ is $+8.44$ with 95\% percentile CI $[4.21, 12.60]$.

\paragraph{Leave-one-rater-out.}
Refitting the LMM with each rater dropped in turn:

\begin{center}
\small
\setlength{\tabcolsep}{6pt}
\renewcommand{\arraystretch}{1.05}
\begin{tabular}{@{}lrl@{}}
\toprule
\textbf{Dropped} & \textbf{coef} & \textbf{95\% CI} \\
\midrule
GPT-5.1 & $+9.05$ & $[4.88, 13.22]$ \\
Claude  & $+8.46$ & $[4.38, 12.53]$ \\
Gemini  & $+8.91$ & $[4.95, 12.87]$ \\
\bottomrule
\end{tabular}
\end{center}

All three $p$-values are $<10^{-4}$.
The effect is consistent across specifications and is not driven by any single rater.

\section{Autoresearch Loop Trajectory}
\label{app:trajectory}

\begin{figure*}[!htbp]
\centering
\includegraphics[width=\textwidth]{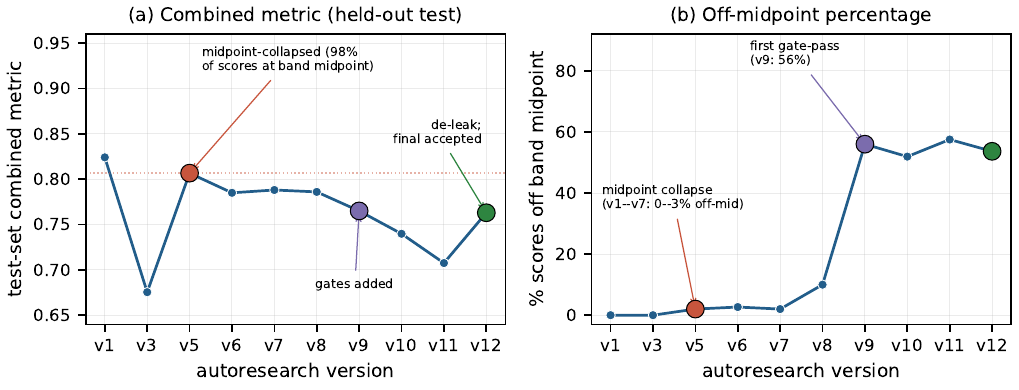}
\caption{Autoresearch loop trajectory across twelve versions.
(a) Held-out test combined metric per version (best accepted prompt; v9--v12 are 2--3-seed means).
The headline figure declines from v5 (0.81, midpoint-collapsed) to v12 (0.76, methodologically honest); the dotted red line marks the v5 peak.
(b) Off-midpoint percentage, fraction of continuous scores not equal to the assigned band's midpoint, averaged across the three raters; computed directly from each version's held-out classifications.
All pre-v8 prompts cluster near 0--3\%, v8's mandatory cue audit lifted this to 10\%, and v9 onward operationalise off-midpoint as a validity gate.
v1 used a mid-tier model lineup; v3 onward use the frontier lineup reported in §\ref{sec:method}.}
\label{fig:trajectory}
\end{figure*}

Figure~\ref{fig:trajectory} plots the held-out test trajectory of the autoresearch loop across the twelve versions that preceded the final accepted prompt.
Panel (a) shows the combined metric; panel (b) shows the off-midpoint percentage (mean across raters), which was first measured at v8 and operationalised as a metric-validity gate at v9.

Two methodologically important transitions are visible.
\textbf{v5} (combined~$=$~0.81) achieved its high agreement through near-universal midpoint scoring (98\% of continuous scores at the assigned band's midpoint), the construct-validity artifact shown in Figure~\ref{fig:midpoint}.
\textbf{v9} introduced the three metric-validity gates (residual ICC, off-midpoint percentage, calibration) described in §\ref{sec:metric}; combined dropped because the loop began penalising midpoint pinning, but off-midpoint percentage rose from $\sim$10\% to 56\%.
\textbf{v12} applied the de-leak step (removed the ``default = Weak Realist'' instruction and the band-frequency priors) and accepted the \texttt{wr\_rs\_boundary} edit, yielding the final test combined $=$ 0.763 with off-midpoint $=$ 53.7\%.

Three model-tier changes are not visible in the figure but affect interpretation.
v1 used a mid-tier lineup (GPT-5.1 + Claude Haiku + Gemini 3 Flash); v3 and later use the frontier lineup reported in §\ref{sec:method} (Claude Sonnet 4.6, GPT-5.1, Gemini 3.0 Pro).
v11 upgraded Gemini to 3.1 Pro Preview.
v4 has no held-out test row and is omitted; its prompt was validated via the v5 baseline run.

\end{document}